\documentclass[pmlr]{jmlr}

\usepackage{longtable}

\usepackage{booktabs}
\usepackage[load-configurations=version-1]{siunitx} 

\makeatletter
\def\set@curr@file#1{\def\@curr@file{#1}} 
\makeatother

\theorembodyfont{\upshape}
\theoremheaderfont{\scshape}
\theorempostheader{:}
\theoremsep{\newline}

\usepackage{enumitem}
\usepackage{algorithm}
\usepackage{graphics}
\usepackage{algorithmic}
\usepackage{multirow}

\title[]{Interleaving Learning, with Application to Neural Architecture Search}

\author{\Name{Hao Ban
}
       \Email{bhsimon0810@gmail.com} 
       \AND
       \Name{Pengtao Xie\textsuperscript{*}}
       \Email{p1xie@eng.ucsd.edu}\\
       \addr 
University of California San Diego
\AND
       }

\begin{document}

\maketitle

\begin{abstract}
Interleaving learning is a human learning technique where a learner interleaves the studies of multiple topics, which  increases long-term retention and improves ability to transfer learned knowledge. Inspired by the interleaving learning technique of humans, in this paper we explore whether this learning methodology is beneficial for improving the performance of machine learning models as well. We propose a novel machine learning framework referred to as interleaving learning (IL). In our framework, a set of  models collaboratively learn a data encoder in an interleaving fashion: the encoder is trained by model 1 for a while, then passed to model 2 for further training, then model 3, and so on; after trained by all models, the encoder returns back to model 1 and is trained again, then moving to model 2, 3, etc. This process repeats for multiple rounds. Our framework is based on multi-level optimization consisting of multiple inter-connected learning stages. An efficient gradient-based algorithm is developed to solve the multi-level optimization problem. We apply interleaving learning to search neural architectures for image classification  on CIFAR-10, CIFAR-100, and ImageNet. The effectiveness of our method is strongly demonstrated by the experimental results. 
\end{abstract}

\section{Introduction}

\let\thefootnote\relax\footnotetext{$^*$Corresponding author.}

\begin{figure}[t]
    \centering
 \includegraphics[width=0.8\columnwidth]{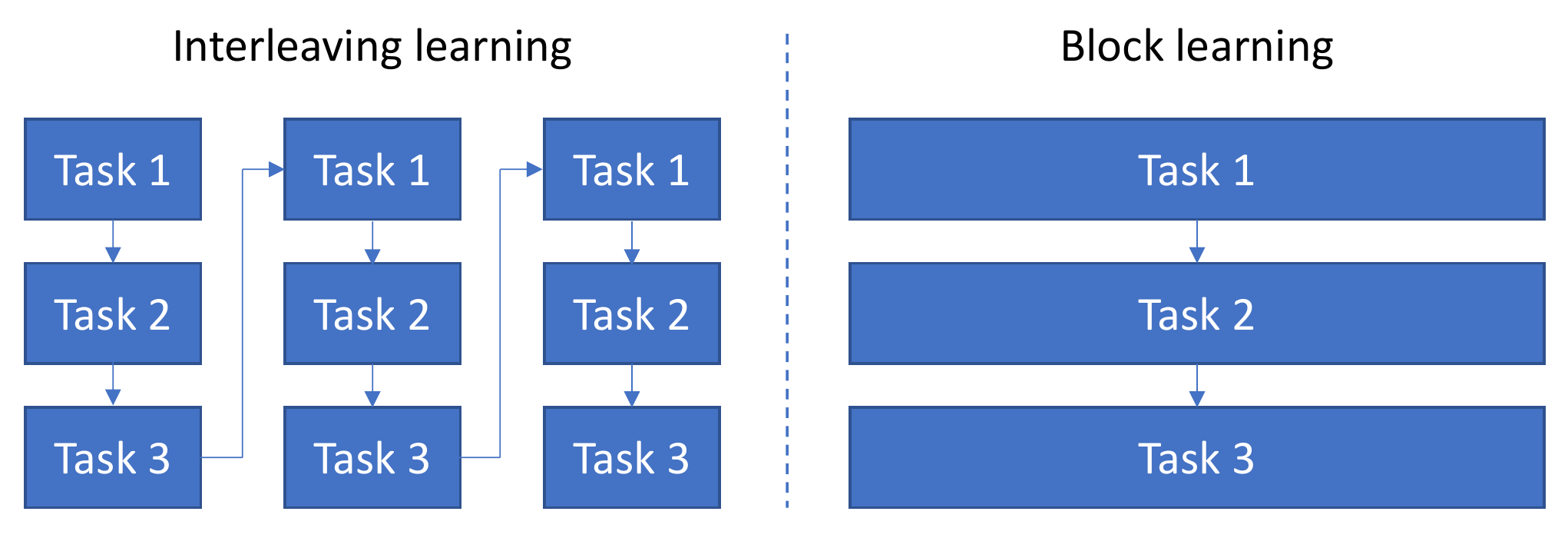}
       \caption{Comparison between  interleaving learning and block learning. In interleaving learning, we perform task 1 for a short while, then move to task 2, then task 3. Afterwards, we move from task 3 back to task 1. This process iterates where each task is performed for a short time period before switching to another task. In contrast, in block learning, we perform task 1 to the very end, then move to task 2, and later task 3. Each task is performed for a long time period until it is completed. Once a task is finished, it will not be performed again. }
 \label{fig:illus}
\end{figure}

Interleaving learning is a learning technique where a learner interleaves the studies of multiple topics: study topic $A$ for a while, then switch to $B$, subsequently to $C$; then switch back to $A$, and so on, forming a pattern of $ABCABCABC\cdots$. Interleaving learning is in contrast to blocked learning, which studies one topic very thoroughly before moving to another topic. Compared with blocked learning, interleaving learning increases long-term retention and improves ability to transfer learned knowledge. Figure~\ref{fig:illus} illustrates the difference between interleaving learning and block learning. 

Motivated by humans' interleaving learning methodology, we are intrigued to explore whether machine learning can be benefited from this learning methodology as well. We propose a novel multi-level optimization framework to formalize the idea of  learning multiple topics in an interleaving way. 
In this framework, we assume there are $K$ learning tasks, each performed by a learner model. Each learner has a data encoder and a task-specific head. The data encoders of all learners share the same architecture, but may have different weight parameters. The $K$ learners perform $M$ rounds of interleaving learning with the following order: 
\begin{equation}
\underbrace{l_1,l_2,\cdots,l_K}_{\textrm{Round }1}
\underbrace{l_1,l_2,\cdots,l_K}_{\textrm{Round }2}
\cdots
\underbrace{l_1,l_2,\cdots,l_K}_{\textrm{Round }m}
\cdots
\underbrace{l_1,l_2,\cdots,l_K}_{\textrm{Round }M}
\end{equation}
where $l_k$ denotes that the $k$-th learner performs learning. In the first round, we first learn $l_1$, then learn $l_2$, and so on. At the end of the first round, $l_K$ is learned. Then we move to the second round, which starts with learning $l_1$, then learns $l_2$, and so on. This pattern repeats until the $M$ rounds of learning are finished. Between two consecutive learners $l_kl_{k+1}$, the encoder weights of the latter learner $l_{k+1}$ are encouraged to be close to the optimally learned encoder weights of the former learner $l_k$. In the interleaving process, the $K$ learners help each other to learn better. Each learner transfers the knowledge learned in its task to the next learner by using its trained encoder to initialize the encoder of the next learner. Meanwhile, each learner leverages the knowledge shared by the previous learner to better train its own model. Via knowledge sharing, in one round of learning, $l_1$ helps $l_2$ to learn better, $l_2$ helps $l_3$ to learn better, and so on. Then moving into the next round, $l_K$ learned in the previous round helps $l_1$ to re-learn for achieving a better learning outcome, then a better $l_1$ further helps $l_2$ to learn better, and so on. After $M$ rounds of learning, each learner uses its model trained in the final round to make predictions on a validation dataset and updates their shared encoder architecture by minimizing the validation losses. Our interleaving learning framework is applied to search neural architectures for image classification on CIFAR-10, CIFAR-100, and ImageNet, where experimental results demonstrate the effectiveness of our method.

The major contributions of this paper are as follows:
\begin{itemize}
\item Drawing insights from a human learning methodology -- interleaving learning, we propose a novel machine learning framework which 
enables  a set of  models to cooperatively train a data encoder in an interleaving way: model 1 trains this encoder for a short time, then hands it over to model 2 to continue the training, then to model 3, etc. When the encoder is trained by all models in one pass, it returns to model 1 and starts the second round of training sequentially by each model. 
This cyclic training process iterates until convergence. During the interleaving process, each model transfers its knowledge to the next model and leverages the knowledge shared by the previous model to learn better. 
\item We formulate interleaving machine learning as a multi-level optimization problem. 
\item We develop an efficient differentiable algorithm  to solve the interleaving learning problem.  
\item We utilize our interleaving learning framework for neural architecture search  on CIFAR-100, CIFAR-10, and ImageNet. Experimental results strongly demonstrate the effectiveness of our method. 
\end{itemize}

The rest of the paper is organized as follows.  Section 2 reviews related works. Section 3 and 4 present the method and experiments respectively.  Section 5 concludes the paper.

\section{Related Works}
The goal of neural architecture search (NAS) is to automatically identify highly-performing neural architectures that can potentially surpass human-designed ones. NAS research has made considerable progress in the past few years.  
Early NAS~\citep{zoph2016neural,pham2018efficient,zoph2018learning} approaches are based on reinforcement learning (RL), where a policy network learns to generate high-quality architectures by maximizing the validation accuracy (as reward). These approaches are conceptually simple and can flexibly perform search in any search spaces. However, they are computationally very demanding. To calculate the reward of a candidate architecture, this architecture needs to be trained on a training dataset, which is very time-consuming.  
To address this issue, differentiable search methods~\citep{cai2018proxylessnas,liu2018darts,xie2018snas} have been proposed. In these methods, each candidate architecture is a combination of many building blocks. The combination coefficients represent the importance of building blocks. Architecture search amounts to learning these differentiable coefficients, which can be done using differentiable optimization algorithms such as gradient descent, with much higher computational efficiency than RL-based approaches.  Differentiable NAS methods started with DARTS~\citep{liu2018darts} and have been improved rapidly since then. For example, P-DARTS \citep{chen2019progressive} allows the architecture depth to increase progressively during searching. It also performs search space regularization and approximation to improve stability of searching algorithms and reduce search cost.  
In PC-DARTS \citep{abs-1907-05737}, the redundancy of search space exploration is reduced by sampling sub-networks from a super network. It also performs operation search in a subset of channels via bypassing the held-out subset in a shortcut.  
Another paradigm of NAS methods~\citep{liu2017hierarchical,real2019regularized} are based on evolutionary algorithms (EA). In these approaches, architectures are considered as individuals in a population. Each architecture is associated with a fitness score representing how good this architecture is. Architectures with higher fitness scores have higher odds of generating offspring (new architectures), which replace architectures that have low-fitness scores. Similar to RL-based methods, EA-based methods are computationally heavy since evaluating the fitness score of an architecture needs to train this architecture. 
Our proposed interleaving learning framework in principle can be applied to any NAS methods. In our experiments, for simplicity and computational efficiency, we choose to work on differentiable NAS methods.

\section{Method}
In this section, we present the details of the interleaving learning framework. There are $K$ learners. Each learner learns to perform a task. These tasks could be the same, e.g., image classification on CIFAR-10; or different, e.g., image classification on CIFAR-10, image classification on ImageNet~\citep{deng2009imagenet}, object detection on MS-COCO~\citep{coco}, etc. Each learner $k$ has a training dataset $D_k^{(\textrm{tr})}$ and a validation dataset $D_k^{(\textrm{val})}$. 
Each learner has a data encoder and a task-specific head performing the target task. For example, if the task is image classification, the data encoder could be a convolutional neural network extracting visual features of the input images and the task-specific head could be a multi-layer perceptron which takes the visual features of an image extracted by the data encoder as input and predicts the class label of this image. We assume the architecture of the data encoder in each learner is learnable. The data encoders of all learners share the same architecture, but their weight parameters could be different in different learners. The architectures of task-specific heads are manually designed by humans  and they could be different in different learners. The $K$ learners perform $M$ rounds of interleaving learning with the following order: 
\begin{equation}
\underbrace{l_1,l_2,\cdots,l_K}_{\textrm{Round }1}
\underbrace{l_1,l_2,\cdots,l_K}_{\textrm{Round }2}
\cdots
\underbrace{l_1,l_2,\cdots,l_K}_{\textrm{Round }m}
\cdots
\underbrace{l_1,l_2,\cdots,l_K}_{\textrm{Round }M}
\end{equation}
where $l_k$ denotes that the $k$-th learner performs learning. In the first round, we first learn $l_1$, then learn $l_2$, and so on. At the end of the first round, $l_K$ is learned. Then we move to the second round, which starts with learning $l_1$, then learns $l_2$, and so on. This pattern repeats until the $M$ rounds of learning are finished. Between two consecutive learners $l_kl_{k+1}$, the weight parameters of the latter learner $l_{k+1}$ are encouraged to be close to the optimally learned encoder weights of the former learner $l_k$. For each learner, the architecture of its encoder remains the same across all rounds; the network weights of the encoder and head can be different in different rounds.

\begin{table}[t]
\caption{Notations in interleaving learning}
\centering
\begin{tabular}{l|p{13cm}}
\hline
Notation & Meaning \\
\hline
$K$ & Number of learners\\
$M$ & Number of rounds\\
$D_k^{(\textrm{tr})}$ & Training dataset of the $k$-th learner\\
$D_k^{(\textrm{val})}$ & Validation dataset of the $k$-th learner\\
$A$ & Encoder architecture shared by all learners\\
$W_k^{(m)}$ & Weight parameters in the data encoder of the $k$-th learner in the $m$-th round\\
$H_k^{(m)}$ & Weight parameters in the task-specific head of the $k$-th learner in the $m$-th round\\
$\widetilde{W}_k^{(m)}$ & The optimal  encoder weights of the $k$-th learner in the $m$-th round\\
$\widetilde{H}_k^{(m)}$ & The optimal weight parameters of the task-specific  head  in the $k$-th learner in the $m$-th round\\
$\lambda$ & Tradeoff parameter\\
\hline
\end{tabular}
\label{tb:notations}
\end{table}

Each learner $k$ has the following learnable parameter sets: 1) architecture $A$ of the encoder; 2) in each round $m$, the learner's encoder has a set of weight parameters $W_k^{(m)}$ specific to this round; 3) in each round $m$, the learner's task-specific head  has a set of weight parameters $H_k^{(m)}$ specific to this round. The encoders of all learners share the same architecture and this architecture remains the same in different rounds. The encoders of different learners have different weight parameters. The weight parameters of a learner's encoder are different in different rounds. Different learners have different task-specific  heads in terms of both architectures and weight parameters. In the interleaving process, the learning of the $k$-th learner is assisted by  the $(k-1)$-th learner. Specifically, during learning, the encoder weights $W_{k}$ of the $k$-th learner are encouraged to be close to the optimal encoder weights $\widetilde{W}_{k-1}$ of the  $(k-1)$-th learner. This is achieved by minimizing the following regularizer : $\|W_{k}-\widetilde{W}_{k-1}\|_2^2$.

There are $M\times K$ learning stages: in each of the $M$ rounds, each of the $K$ learners is learned in a stage. In the very first learning stage, the first learner in the first round is learned. It trains the weight parameters of its data encoder and the weight parameters of its task-specific head on its training dataset.  The optimization problem is:
\begin{equation}
\widetilde{W}_1^{(1)}(A) =\textrm{min}_{W^{(1)}_1,H^{(1)}_1} \; L(A, W^{(1)}_1, H^{(1)}_1,D_1^{(\textrm{tr})}).
\end{equation}
In this optimization problem, $A$ is not learned. Otherwise, a trivial solution of $A$ will be resulted in. In this trivial solution, $A$ would be excessively large and expressive, and can perfectly overfit the training data, but will have poor generalization capability on unseen data. After learning, the optimal head is discarded. The optimal encoder weights $\widetilde{W}_1^{(1)}(A)$ are a function of $A$ since the training loss is a function of $A$ and $\widetilde{W}_1$ is a function of the training loss.  $\widetilde{W}_1^{(1)}(A)$ is passed to the next learning stage to help with the learning of the second learner. 

In any other learning stage, e.g., the $l$-th stage where the learner is $k$ and the round of interleaving is $m$, the optimization problem is:
\begin{equation*}
\begin{array}{l}
   \widetilde{W}_k^{(m)}(A)=
   \underset{W_k^{(m)},H_k^{(m)}}{\textrm{min}}
\; L(A, W_k^{(m)}, H_k^{(m)},D_k^{(\textrm{tr})})+\lambda\|W^{(m)}_k-\widetilde{W}_{l-1}(A)\|^2_{2},
\end{array}
\label{eq:reg}
\end{equation*}
where $\|W^{(m)}_k-\widetilde{W}_{l-1}\|_2^2$ encourages the encoder weights $W^{(m)}_k$ at this stage to be close to the optimal encoder weights $\widetilde{W}_{l-1}$ learned in the previous stage and $\lambda$ is a tradeoff parameter. 
The optimal encoder weights $\widetilde{W}_k^{(m)}(A)$ are a function of the encoder architecture $A$. The encoder architecture is not updated at this learning stage, for the same reason described above. In the round of 1 to $M-1$, the optimal heads are discarded after learning. In the round of $M$, the optimal heads $\{\widetilde{H}_k^{(M)}(A)\}_{k=1}^K$ are retained and will be used in the final learning stage. In the final stage, each learner evaluates its model learned in the final round $M$ on the validation set. The encoder architecture $A$ is learned by minimizing the validation losses of all learners. The corresponding optimization problem is:
\begin{equation}
    \textrm{min}_{A} \;
  \sum_{k=1}^K L(A, \widetilde{W}_k^{(M)}(A), \widetilde{H}_k^{(M)}(A),D_k^{(\textrm{val})}).  
\end{equation}

To this end, we are ready to formulate the interleaving learning problem using a multi-level optimization framework, as shown in Eq.(\ref{eq:il}). From bottom to top, the $K$ learners perform $M$ rounds of interleaving learning. Learners in adjacent learning stages are coupled via $\|W_{k}-\widetilde{W}_{k-1}\|_2^2$. The architecture $A$ is  learned by minimizing the validation loss. Similar to~\citep{liu2018darts}, we represent $A$ in a differentiable way. $A$ is a weighted combination of multiple layers of basic building blocks such as convolution, pooling, normalization, etc. The output of each building block is multiplied with a weight indicating how important this block is. During architecture search, these differentiable weights are learned. After the search process, blocks with large weights are retained to form the final architecture.

\begin{equation}
\begin{array}{ll}
   \textrm{min}_{A}  & 
  \sum_{k=1}^K L(A, \widetilde{W}_k^{(M)}(A), \widetilde{H}_k^{(M)}(A),D_k^{(\textrm{val})}) 
   \\
      s.t.  
        & \textrm{\textbf{Round $\mathbf{M}$}:}\\
      & \widetilde{W}_K^{(M)}(A),\widetilde{H}_K^{(M)}(A)=\textrm{min}_{W_K^{(M)},H^{(M)}_K} \quad L(A, W_K^{(M)}, H^{(M)}_K,D_K^{(\textrm{tr})})+\lambda\|W^{(M)}_K-\widetilde{W}_{K-1}^{(M)}(A)\|^2_{2}\\
      & \cdots \\
    & \widetilde{W}_1^{(M)}(A), \widetilde{H}_1^{(M)}(A)=\textrm{min}_{W^{(M)}_1,H^{(M)}_1} \quad L(A, W^{(M)}_1, H^{(M)}_1,D_1^{(\textrm{tr})})+\lambda\|W^{(M)}_1-\widetilde{W}_K^{(M-1)}(A)\|^2_{2}\\
    & \cdots\\
        & \textrm{\textbf{Round 2}:}\\
      & \widetilde{W}_K^{(2)}(A) =\textrm{min}_{W_K^{(2)},H^{(2)}_K} \quad L(A, W_K^{(2)}, H^{(2)}_K,D_K^{(\textrm{tr})})+\lambda\|W^{(2)}_K-\widetilde{W}_{K-1}^{(2)}(A)\|^2_{2}\\
      & \cdots \\
    & \widetilde{W}_1^{(2)}(A) =\textrm{min}_{W^{(2)}_1,H^{(2)}_1} \quad L(A, W^{(2)}_1, H^{(2)}_1,D_1^{(\textrm{tr})})+\lambda\|W^{(2)}_1-\widetilde{W}_K^{(1)}(A)\|^2_{2}\\
   & \textrm{\textbf{Round 1}:}\\
      & \widetilde{W}_K^{(1)}(A) =\textrm{min}_{W_K^{(1)},H^{(1)}_K} \quad L(A, W_K^{(1)}, H^{(1)}_K,D_K^{(\textrm{tr})})+\lambda\|W^{(1)}_K-\widetilde{W}_{K-1}^{(1)}(A)\|^2_{2}\\
      & \cdots \\
      & \widetilde{W}_k^{(1)}(A) =\textrm{min}_{W_k^{(1)},H^{(1)}_k} \quad L(A, W_k^{(1)}, H^{(1)}_k,D_k^{(\textrm{tr})})+\lambda\|W^{(1)}_k-\widetilde{W}_{k-1}^{(1)}(A)\|^2_{2}\\
      & \cdots \\
      & \widetilde{W}_2^{(1)}(A) =\textrm{min}_{W_2^{(1)},H^{(1)}_2} \quad L(A, W^{(1)}_2, H^{(1)}_2,D_2^{(\textrm{tr})})+\lambda\|W^{(1)}_2-\widetilde{W}_1^{(1)}(A)\|^2_{2}\\
    & \widetilde{W}_1^{(1)}(A) =\textrm{min}_{W^{(1)}_1,H^{(1)}_1} \quad L(A, W^{(1)}_1, H^{(1)}_1,D_1^{(\textrm{tr})})
\end{array}
\label{eq:il}
\end{equation}

\begin{algorithm}[H]
\SetAlgoLined
 \While{not converged}{
1. Update $\widetilde{W}_1^{(1)}(A)$ using Eq.(\ref{eq:w11})\\
2. For $k=2\cdots K$, update $\widetilde{W}_k^{(1)}(A)$ using Eq.(\ref{eq:wkm})\\
3. For $k=1\cdots K$ and $m=2\cdots M$, update $\widetilde{W}_k^{(m)}(A)$ using Eq.(\ref{eq:wkm})\\
4. For $k=1\cdots K$, update $\widetilde{H}_k^{(M)}(A)$ using Eq.(\ref{eq:hm})\\
5. Update $A$ using Eq.(\ref{eq:a-il})
 }
 \caption{Optimization algorithm for interleaving learning}
 \label{algo:algo-il}
\end{algorithm}
\subsection{Optimization Algorithm}

In this section,  we develop an optimization algorithm for interleaving learning. For each optimization problem $\widetilde{W}_k^{(m)}(A) =\textrm{min}_{W^{(m)}_k,H^{(m)}_k} \quad L(A, W^{(m)}_k, H^{(m)}_k,D_k^{(\textrm{tr})})+\lambda\|W^{(m)}_k-\widetilde{W}_{k-1}^{(m)}(A)\|^2_{2}
$ in a learning stage, we approximate the optimal solution $\widetilde{W}_k^{(m)}(A)$ by one-step gradient descent update of the optimization variable $W^{(m)}_k$: 
\begin{equation*}
\begin{array}{l}
    \widetilde{W}_k^{(m)}(A)\approx \overline{W}_k^{(m)}(A)= W^{(m)}_k-\eta \nabla_{W^{(m)}_k}( L(A, W^{(m)}_k, H^{(m)}_k,D_k^{(\textrm{tr})})+\lambda\|W^{(m)}_k-\widetilde{W}_{k-1}^{(m)}(A)\|^2_{2}).
    \end{array}
\end{equation*}
For $\widetilde{W}_1^{(1)}(A)$, the approximation is:
\begin{equation}
    \widetilde{W}_1^{(1)}(A)\approx  \overline{W}_1^{(1)}(A)= W^{(1)}_1-\eta \nabla_{W^{(1)}_1} L(A, W^{(1)}_1, H^{(1)}_1,D_1^{(\textrm{tr})}).
    \label{eq:w11}
\end{equation}
For $\widetilde{W}_k^{(m)}(A)$, the approximation is:
\begin{equation}
\begin{array}{l}
    \overline{W}_k^{(m)}(A)= W^{(m)}_k-\eta \nabla_{W^{(m)}_k} L(A, W^{(m)}_k, H^{(m)}_k,D_k^{(\textrm{tr})})-2\eta\lambda(W^{(m)}_k-\overline{W}_{k-1}^{(m)}(A)),
    \label{eq:wkm}
    \end{array}
\end{equation}
where $\overline{W}_{k-1}^{(m)}(A)$ is the approximation of $\widetilde{W}_{k-1}^{(m)}(A)$. Note that $\{\overline{W}_k^{(m)}(A)\}_{k,m=1}^{K,M}$ are  calculated recursively, where $\overline{W}_k^{(m)}(A)$ is a function of $\overline{W}_{k-1}^{(m)}(A)$, $\overline{W}_{k-1}^{(m)}(A)$ is a function of $\overline{W}_{k-2}^{(m)}(A)$, and so on. When $m>1$ and $k=1$,  $\overline{W}_{k-1}^{(m)}(A)=\overline{W}_{K}^{(m-1)}(A)$. For $\widetilde{H}_k^{(M)}(A)$, the approximation is:
\begin{equation}
     \overline{H}_k^{(M)}(A)=H_k^{(M)}(A)-\eta \nabla_{H_k^{(M)}(A)}L(A, W_k^{(M)}, H^{(M)}_k,D_k^{(\textrm{tr})}).
    \label{eq:hm}
\end{equation}
In the validation stage, we plug the approximations of $\{\widetilde{W}_k^{(M)}(A)\}_{k=1}^K$ and $\{\widetilde{H}_k^{(M)}(A)\}_{k=1}^K$ into the validation loss function, calculate the gradient of the approximated objective w.r.t the encoder architecture $A$, then update $A$ via:
\begin{equation}
    A\gets A-\eta \sum_{k=1}^K \nabla_AL(A, \overline{W}_k^{(M)}(A), \overline{H}_k^{(M)}(A),D_k^{(\textrm{val})}).
    \label{eq:a-il}
\end{equation}
The update steps from Eq.(\ref{eq:w11}) to Eq.(\ref{eq:a-il}) iterate until convergence. The entire algorithm is summarized in Algorithm~\ref{algo:algo-il}.

\section{Experiments}
In this section, we apply the proposed interleaving ML framework for neural architecture search in image classification tasks. Following the experimental protocol in \citep{liu2018darts}, each experiment consists of an architecture search phrase and an architecture evaluation phrase. In the search phrase, an optimal architecture cell is searched by minimizing the validation loss. In the evaluation phrase, a larger network is created by stacking multiple copies of the optimally searched cell. This new network is re-trained from scratch and evaluated on the test set.

\subsection{Datasets}
\label{sec:datasets}
Three popular image classification datasets are involved in the experiments: CIFAR-10, CIFAR-100,  and ImageNet~\citep{deng2009imagenet}. CIFAR-10 contains 60K images from 10 classes. CIFAR-100 contains 60K images from 100 classes. ImageNet contains 1.25 million images from 1000 classes. For CIFAR-10 and CIFAR-100, each of them is split into train/validation/test sets with 25K/25K/10K images respectively. For ImageNet, it has 1.2M training images and 50K test images.

\subsection{Baselines}
Our IL framework can be generally used together with any differentiable NAS method. In the experiments, we apply IL to three widely-used NAS methods: DARTS~\citep{liu2018darts}, P-DARTS~\citep{chen2019progressive}, and PC-DARTS~\citep{abs-1907-05737}. The search space of these methods are similar, where the building blocks include $3\times 3$ and $5\times 5$ (dilated) separable convolutions, $3\times 3$ max pooling, $3\times 3$ average pooling, identity, and zero. We compare our interleaving framework with a multi-task learning framework where a shared encoder architecture is searched simultaneously on CIFAR-10 and CIFAR-100. 
The formulation is:
\begin{equation}
\begin{array}{ll}
   \textrm{min}_{A}  & 
   L(A, \widetilde{W}_{100}(A), \widetilde{H}_{100}(A),D_{100}^{(\textrm{val})})+    \alpha L(A, \widetilde{W}_{10}(A), \widetilde{H}_{10}(A),D_{10}^{(\textrm{val})})
   \\
      s.t.  
      & \widetilde{W}_{100}(A), \widetilde{H}_{100}(A), \widetilde{W}_{10}(A), \widetilde{H}_{10}(A)=\\
      &\textrm{min}_{W_{100},H_{100},W_{10},H_{10}} \quad L(A, W_{100}, H_{100},D_{100}^{(\textrm{tr})})  + \beta L(A, W_{10}, H_{10},D_{10}^{(\textrm{tr})})
\end{array}
\label{eq:ab-il}
\end{equation}
where $W_{100}$ and $H_{100}$ are the encoder weights and classification head for CIFAR-100. $W_{10}$ and $H_{10}$ are the encoder weights and classification head for CIFAR-10. $D_{100}^{(\textrm{tr})}$ and $D_{100}^{(\textrm{val})}$ are the training and validation sets of CIFAR-100. $D_{10}^{(\textrm{tr})}$ and $D_{10}^{(\textrm{val})}$ are the training and validation sets of CIFAR-10. $A$ is the encoder architecture shared by CIFAR-100 and CIFAR-10. $\alpha$ and $\beta$ in Eq.(\ref{eq:ab-il}) are both set to 1.

\subsection{Experimental Settings}
\label{sec:settings}

In the interleaving learning framework, we set two learners: one learns to classify CIFAR-10 images and the other learns to classify CIFAR-100 images. Each learner has an image encoder and a classification head. Encoders of these two learners share the same architecture, whose search space is the same as that in DARTS/P-DARTS/PC-DARTS. The encoder is a stack of 8 cells, each consisting of 7 nodes. The initial channel number was set to 16. 
For the learner on CIFAR-10, the classification head is a 10-way linear classifier. The training and validation set of CIFAR-10 is used as $D_1^{(\textrm{tr})}$ and $D_1^{(\textrm{val})}$ respectively. For the learner on CIFAR-100, the classification head is a 100-way linear classifier. The training and validation set of CIFAR-100 is used as $D_2^{(\textrm{tr})}$ and $D_2^{(\textrm{val})}$ respectively. We set the number of interleaving rounds to 2. The tradeoff parameter $\lambda$ in Eq.(\ref{eq:il}) is set to 100. The order of tasks in the interleaving process is: CIFAR-100, CIFAR-10, CIFAR-100, CIFAR-10. 

During architecture search, network weights were optimized using the SGD optimizer with a batch size of 64, an initial learning rate of 0.025, a learning rate scheduler of cosine decay, a weight decay of 3e-4,  a momentum of 0.9, and an epoch number of 50. The architecture variables were optimized using the Adam~\citep{adam} optimizer with a learning rate of 3e-4 and a weight decay of 1e-3. The rest hyperparameters follow those in DARTS, P-DARTS, and PC-DARTS. 

Given the optimally searched architecture cell, we evaluate it individually on CIFAR-10, CIFAR-100, and ImageNet. 
For CIFAR-10 and CIFAR-100, we stack 20 copies of the searched cell into a larger network as the image encoder. The initial channel number was set to 36. We trained the network for 600 epochs on the combination of the training and validation datasets  where the mini-batch size was set to 96. The experiments were conducted on one Tesla v100 GPU.  
For ImageNet, similar to~\citep{liu2018darts}, we evaluate the architecture cells searched on CIFAR10/100. A larger network is formed by stacking 14 copies of the searched cell. The  initial channel number was set to 48. We trained the network for 250 epochs on the 1.2M training images using eight Tesla v100 GPUs where the batch size was set to 1024. 
Each IL experiment was repeated for ten times with different random initialization.  Mean and standard deviation of classification errors obtained from the 10 runs are reported.

\subsection{Results}

\begin{table}[t]
\caption{Classification errors on the test set of CIFAR-100, number of model parameters, and search cost (GPU days). 
    IL-DARTS-2nd denotes that our proposed interleaving learning (IL) framework is applied to the search space of DARTS-2nd.   DARTS-1st  and 
DARTS-2nd means that first order and second order approximation is used in DARTS' optimization procedure. Results marked with *
    are taken from DARTS$^{-}$ \citep{abs-2009-01027}. Methods marked with $\dag$ were re-run for 10 times with different random initialization. $\Delta$ denotes this algorithm ran for 600 epochs instead of 2000 epochs in the architecture evaluation stage, to ensure 
    the comparison with other methods (which all ran for 600 epochs) is fair. Search cost is measured by GPU days on a Tesla v100.
    }
    \centering
    \begin{tabular}{l|ccc}
    \toprule
    Method & Error(\%)& Param(M)& Cost\\
    \midrule
    *ResNet \citep{he2016deep}&22.10&1.7&-\\
     *DenseNet \citep{HuangLMW17}&17.18&25.6 &-\\
    \hline
    *PNAS \citep{LiuZNSHLFYHM18}&19.53&3.2&150\\
    *ENAS \citep{pham2018efficient}&19.43&4.6&0.5\\
        *AmoebaNet \citep{real2019regularized}&18.93&3.1&3150\\
    \hline
               ${}^{\dag}$DARTS-1st \citep{liu2018darts}  &20.52$\pm$0.31 &1.8 &0.4\\
    *GDAS \citep{DongY19}&18.38&3.4&0.2\\
    *R-DARTS \citep{ZelaESMBH20}&18.01$\pm$0.26&-&1.6
    \\
               *DARTS$^{-}$ \citep{abs-2009-01027}&17.51$\pm$0.25&3.3&0.4\\      ${}^{\dag}$DARTS$^{-}$ \citep{abs-2009-01027}& 18.97$\pm$0.16& 3.1&0.4\\
  ${}^{\Delta}$DARTS$^{+}$ \citep{abs-1909-06035}&17.11$\pm$0.43&3.8&0.2\\
      *DropNAS \citep{HongL0TWL020} & 16.39&4.4&0.7 \\
\hline
\hline
            *DARTS-2nd \citep{liu2018darts}  & 20.58$\pm$0.44&1.8&1.5 \\
            $\;\;$MTL-DARTS2nd & 18.92$\pm$0.17 & 2.4& 3.1 \\
             $\;\;$IL-DARTS2nd (ours) & \textbf{17.12}$\pm$0.08 & 2.6& 3.2 \\
             \hline              *P-DARTS \citep{chen2019progressive}&17.49&3.6&0.3\\
             $\;\;$MTL-PDARTS &17.67$\pm$0.31 &3.5 &0.6\\
             $\;\;$IL-PDARTS (ours)& \textbf{16.14}$\pm$0.17& 3.6&0.6\\
             \hline
                              $\dag$PC-DARTS \citep{abs-1907-05737} &17.96$\pm$0.15&3.9&0.1 \\
                        $\;\;$MTL-PCDARTS & 18.11$\pm$0.27& 3.9&0.2\\
                      $\;\;$IL-PCDARTS (ours)&17.83$\pm$0.14 &3.8 &0.3\\
        \bottomrule
    \end{tabular}
    \label{tab:cifar100-il}
\end{table}

\begin{table}[t]
\caption{
    Classification errors on the test set of CIFAR-10, number of model parameters, and search cost. Results marked with * are taken from  DARTS$^{-}$ \citep{abs-2009-01027}, NoisyDARTS \citep{abs-2005-03566},  and DrNAS \citep{abs-2006-10355}.
    The rest notations are the same as those in Table~\ref{tab:cifar100-il}.
   }
    \centering
    \begin{tabular}{l|ccc}
    \toprule
    Method& Error(\%)& Param(M) & Cost\\
    \midrule
    *DenseNet
    \citep{HuangLMW17}&3.46&25.6 &-\\
    \hline
     *HierEvol \citep{liu2017hierarchical}&3.75$\pm$0.12& 15.7 &300\\
    *NAONet-WS \citep{LuoTQCL18} & 3.53 & 3.1&0.4 \\
        *PNAS \citep{LiuZNSHLFYHM18} &3.41$\pm$0.09  &3.2& 225\\
        *ENAS \citep{pham2018efficient} &2.89 & 4.6  &0.5 \\
    *NASNet-A \citep{zoph2018learning} & 2.65 & 3.3& 1800\\
    *AmoebaNet-B \citep{real2019regularized} & 2.55$\pm$0.05 & 2.8&3150  \\
    \hline
                *DARTS-1st \citep{liu2018darts} &3.00$\pm$0.14&3.3&  0.4\\
        *R-DARTS \citep{ZelaESMBH20} &2.95$\pm$0.21  &- & 1.6 \\
            *GDAS \citep{DongY19}&2.93& 3.4& 0.2 \\
    *SNAS \citep{xie2018snas} &2.85 & 2.8& 1.5\\
    ${}^{\Delta}$DARTS$^{+}$ \citep{abs-1909-06035}&2.83$\pm$0.05&3.7&0.4\\
        *BayesNAS \citep{ZhouYWP19} &2.81$\pm$0.04 &3.4&0.2 \\
        *MergeNAS \citep{WangXYYHS20} &2.73$\pm$0.02 &2.9 & 0.2 \\
        *NoisyDARTS \citep{abs-2005-03566} &2.70$\pm$0.23&3.3  & 0.4 \\
            *ASAP \citep{NoyNRZDFGZ20} &2.68$\pm$0.11 & 2.5&0.2 \\
                *SDARTS
    \citep{abs-2002-05283}&2.61$\pm$0.02 & 3.3& 1.3 \\
         *DARTS$^{-}$ \citep{abs-2009-01027}&2.59$\pm$0.08&  3.5&0.4\\
         ${}^{\dag}$DARTS$^{-}$ \citep{abs-2009-01027}& 2.97$\pm$0.04& 3.3&0.4\\
            *DropNAS \citep{HongL0TWL020} &2.58$\pm$0.14 & 4.1&0.6 \\
    *FairDARTS \citep{abs-1911-12126} &2.54 &3.3 &0.4 \\
        *DrNAS \citep{abs-2006-10355} &2.54$\pm$0.03&4.0&  0.4\\
     \hline
        \hline
               *DARTS-2nd \citep{liu2018darts} &2.76$\pm$0.09&3.3&  1.5\\ $\;\;$MTL-DARTS2nd   & 2.91$\pm$0.12  &2.4 & 3.1 \\
           $\;\;$IL-DARTS2nd (ours)  & \textbf{2.62}$\pm$0.04 &2.6 & 3.2\\
           \hline
                    *PC-DARTS \citep{abs-1907-05737} &2.57$\pm$0.07&3.6& 0.1 \\
           $\;\;$MTL-PCDARTS  &2.63$\pm$0.05&3.9& 0.2\\
          $\;\;$IL-PCDARTS (ours) &2.55$\pm$0.11&3.8&0.3 \\
          \hline 
                     *P-DARTS \citep{chen2019progressive} &2.50 &3.4&0.3\\ 
                      $\;\;$MTL-PDARTS  & 2.63$\pm$0.12 & 3.5& 0.6\\
                     $\;\;$IL-PDARTS (ours) & 2.51$\pm$0.10 & 3.6& 0.6\\
        \bottomrule
    \end{tabular}
    \label{tab:c10-il}
\end{table}

\begin{table*}[t]
\caption{Top-1 and top-5 classification errors on the test set of ImageNet, number of model parameters, and search cost (GPU days). Results marked with * were taken from DARTS$^{-}$ \citep{abs-2009-01027} and DrNAS \citep{abs-2006-10355}. 
    The rest notations are the same as those in Table~\ref{tab:cifar100-il}. From top to bottom, on the first, second, and third block are: 1) networks manually designed by humans; 2) non-differentiable architecture search methods; and 3) differentiable search methods. 
    }
    \centering
    \begin{tabular}{l|cccc}
    \toprule
  \multirow{2}{*}{Method}   & Top-1  &Top-5 &Param & Cost \\
         & Error (\%) & Error (\%)&(M) & (GPU days)\\
    \midrule
    *Inception-v1 \citep{googlenet}&30.2 &10.1&6.6&- \\
    *MobileNet \citep{HowardZCKWWAA17} &  29.4& 10.5 &4.2&- \\
    *ShuffleNet 2$\times$ (v1) \citep{ZhangZLS18} &  26.4 &10.2 & 5.4&-\\
    *ShuffleNet 2$\times$ (v2) \citep{MaZZS18} &  25.1 &7.6 & 7.4&-\\
    \hline
    *NASNet-A \citep{zoph2018learning} &26.0 &8.4 &5.3 &1800 \\
    *PNAS \citep{LiuZNSHLFYHM18} &25.8 &8.1  &5.1 &225 \\
    *MnasNet-92 \citep{TanCPVSHL19} & 25.2 & 8.0& 4.4&1667\\
        *AmoebaNet-C \citep{real2019regularized} &  24.3 &7.6 &6.4&3150 \\
    \hline
     *SNAS \citep{xie2018snas} & 27.3 &9.2 &4.3 &1.5 \\
          *BayesNAS \citep{ZhouYWP19} &26.5 &8.9 &3.9&0.2 \\
                    *PARSEC \citep{abs-1902-05116} & 26.0 &8.4&5.6&1.0 \\
     *GDAS \citep{DongY19} &  26.0&8.5 &5.3 & 0.2\\
                 *DSNAS \citep{HuXZLSLL20} &25.7& 8.1 &- & -\\
          *SDARTS-ADV \citep{abs-2002-05283}&25.2& 7.8 &5.4& 1.3 \\
           *PC-DARTS \citep{abs-1907-05737} & 25.1 &7.8&5.3&0.1\\
                *ProxylessNAS \citep{cai2018proxylessnas} & 24.9 &7.5 &7.1 &8.3  \\
          *FairDARTS (CIFAR-10) \citep{abs-1911-12126} &24.9 &7.5 &4.8 &0.4 \\ 
     *FairDARTS (ImageNet) \citep{abs-1911-12126} &24.4 &7.4 &4.3 &3.0 \\
             *DrNAS \citep{abs-2006-10355} & 24.2 &7.3& 5.2&3.9\\ 
         *DARTS$^{+}$ (ImageNet) \citep{abs-1909-06035}& 23.9& 7.4&5.1&6.8\\
        *DARTS$^{-}$ \citep{abs-2009-01027}&23.8& 7.0&4.9&4.5\\
     *DARTS$^{+}$ (CIFAR-100) \citep{abs-1909-06035}&23.7& 7.2&5.1&0.2\\
     \hline
      \hline
         *DARTS2nd-CIFAR10 \citep{liu2018darts}  & 26.7 &8.7&4.7&1.5 \\
           $\;\;$MTL-DARTS2nd-CIFAR10/100 & 26.4& 8.5  &3.5 & 3.1\\
        $\;\;$IL-DARTS2nd-CIFAR10/100 (ours) & \textbf{25.5} &\textbf{8.0} &3.8 & 3.2\\
        \hline
          *PDARTS-CIFAR10 \citep{chen2019progressive}&24.4 &7.4&4.9&0.3\\ 
             *PDARTS-CIFAR100 \citep{chen2019progressive}&24.7& 7.5&5.1&0.3\\
              $\;\;$MTL-PDARTS-CIFAR10/100  & 25.0 & 7.9 &5.1 &0.6 \\
           $\;\;$IL-PDARTS-CIFAR10/100 (ours) & \textbf{24.1} & \textbf{7.1} & 5.3&0.6 \\ 
         \bottomrule
    \end{tabular}
    \label{tab:imagenet-il}
\end{table*}

Table~\ref{tab:cifar100-il} and Table~\ref{tab:c10-il} show the classification errors on the test sets of CIFAR-100 and CIFAR-10 respectively, together with the number of model parameters and search costs (GPU days) of different NAS methods. From these two tables, we make the following observations. \textbf{First}, when our proposed interleaving learning (IL) framework is applied to different differentiable NAS methods, the errors of these methods can be greatly reduced. For example, on CIFAR-100, IL-DARTS2nd (applying IL to DARTS) achieves an average error of 17.12\%, which is significantly lower than the error of vanilla DARTS-2nd, which is 20.58\%. As another example, the error of P-DARTS on CIFAR-100  is 17.49\%; applying IL to  P-DARTS, this error is reduced to 16.14\%. On CIFAR-10, applying IL to DARTS-2nd reduces the error from 2.76\% to 2.62\%. These results demonstrate the effectiveness of interleaving learning. In IL, the encoder trained on CIFAR-100 is used to initialize the encoder for CIFAR-10. Likewise, the encoder trained on CIFAR-10 is used to help with the learning of the encoder on CIFAR-100. These two procedures iterates, which enables the learning tasks on CIFAR-100 and CIFAR-10 to mutually benefit each other. In contrast, in baselines including DARTS-2nd, P-DARTS, and PC-DARTS, the encoders for CIFAR-100 and CIFAR-10 are learned separately without interleaving; there is no mechanism to let the learning on CIFAR-100  benefit the learning on CIFAR-10 and vice versa. Overall, the improvement achieved by our method on CIFAR-100 is more significant than that on CIFAR-10. This is probably because CIFAR-10 is a relatively easy dataset for classification (with 10 classes only), which leaves smaller room for improvement. With 100 classes, CIFAR-100 is more challenging for classification and can better differentiate the capabilities of different methods. 
\textbf{Second}, interleaving learning (IL) performs better than multi-task learning (MTL). For example, on CIFAR-100, when applied to DARTS-2nd, the error of IL is lower than that of MTL; this is also the case when applied to P-DARTS and PC-DARTS. On CIFAR-10, when applied to DARTS-2nd, P-DARTS, and PC-DARTS, IL outperforms MTL as well.  In the inner optimization problem of the MTL formulation, the encoder weights $W_{100}$ for CIFAR-100 and the encoder weights   $W_{10}$ for CIFAR-10 are trained independently without a mechanism of mutually benefiting each other. In contrast, IL enables $W_{100}$ and $W_{10}$ to help each other for better training via the interleaving mechanism. 
\textbf{Third}, among all the methods in Table~\ref{tab:cifar100-il}, our IL-PDARTS method achieves the lowest error, which shows that our IL method is highly competitive in pushing the limit of the state-of-the-art. \textbf{Fourth}, while our IL method achieves better accuracy, it does not substantially increase model size (number of parameters) or search cost. 

Table~\ref{tab:imagenet-il} shows the top-1 and top-5 classification errors on the test set of ImageNet, number of model parameters, and search cost (GPU days). Similar to the observations made from Table~\ref{tab:cifar100-il} and Table~\ref{tab:c10-il}, the results on ImageNet show the following. \textbf{First}, when applying our IL framework to DARTS and P-DARTS, the errors of these methods can be greatly reduced. For example, IL-DARTS2nd-CIFAR10/100 (applying IL to DARTS-2nd and searching the architecture on CIFAR-10 and CIFAR-100) achieves a top-1 error of 25.5\% and top-5 error of 8.0\%; without IL, the top-1 and top-5 error of DARTS2nd-CIFAR10 is 26.7\% and 8.7\%. As another example, the errors achieved by IL-PDARTS-CIFAR10/100 are much lower than those of PDARTS-CIFAR100 and PDARTS-CIFAR10. These results further demonstrate the effectiveness of interleaving learning which enables different tasks to mutually help each other. \textbf{Second}, interleaving learning (IL) outperforms multi-task learning (MTL). For example, IL-DARTS2nd-CIFAR10/100 achieves lower errors than MTL-DARTS2nd-CIFAR10/100; IL-PDARTS-CIFAR10/100 performs better than MTL-PDARTS-CIFAR10/100. These results further show that making different tasks help each other in an interleaving and cyclic way is more advantageous than performing them jointly and simultaneously.
 \textbf{Third}, while our IL framework can greatly improve classification accuracy, it does not increase the parameter number and search cost substantially.

\subsection{Ablation Studies}
We perform ablation studies to check the effectiveness of individual modules in our framework. In each ablation study, the ablation setting is compared with the full interleaving learning framework.

\begin{figure}[t]
    \centering
 \includegraphics[width=0.49\columnwidth]{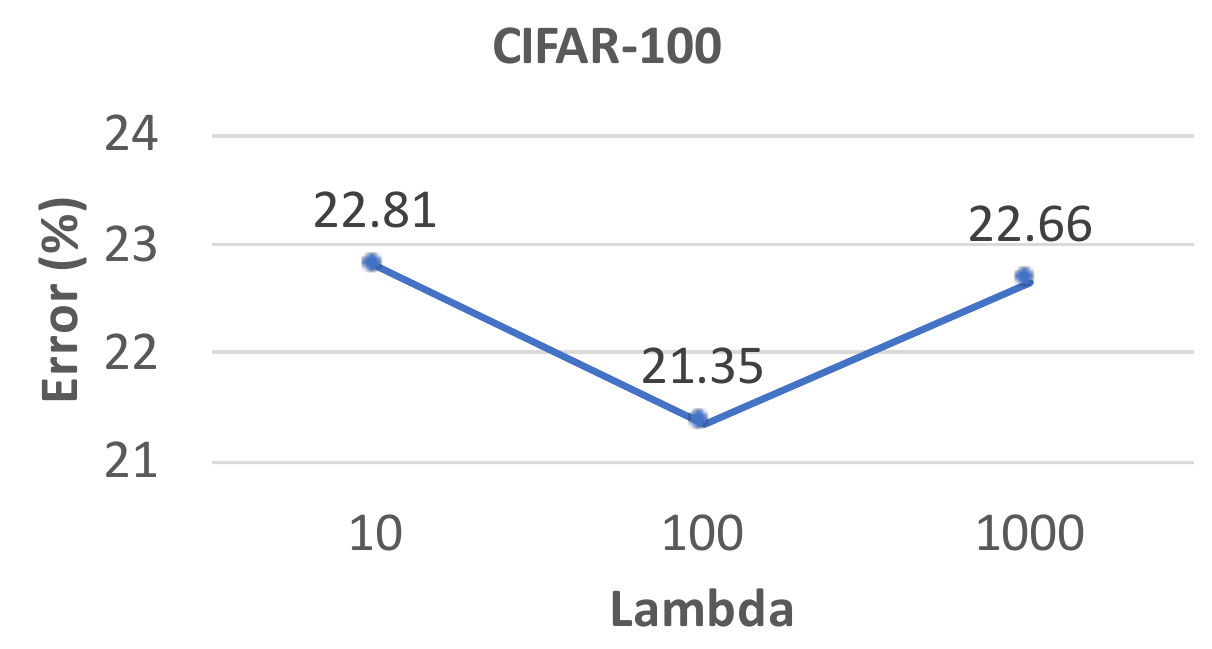}
  \includegraphics[width=0.49\columnwidth]{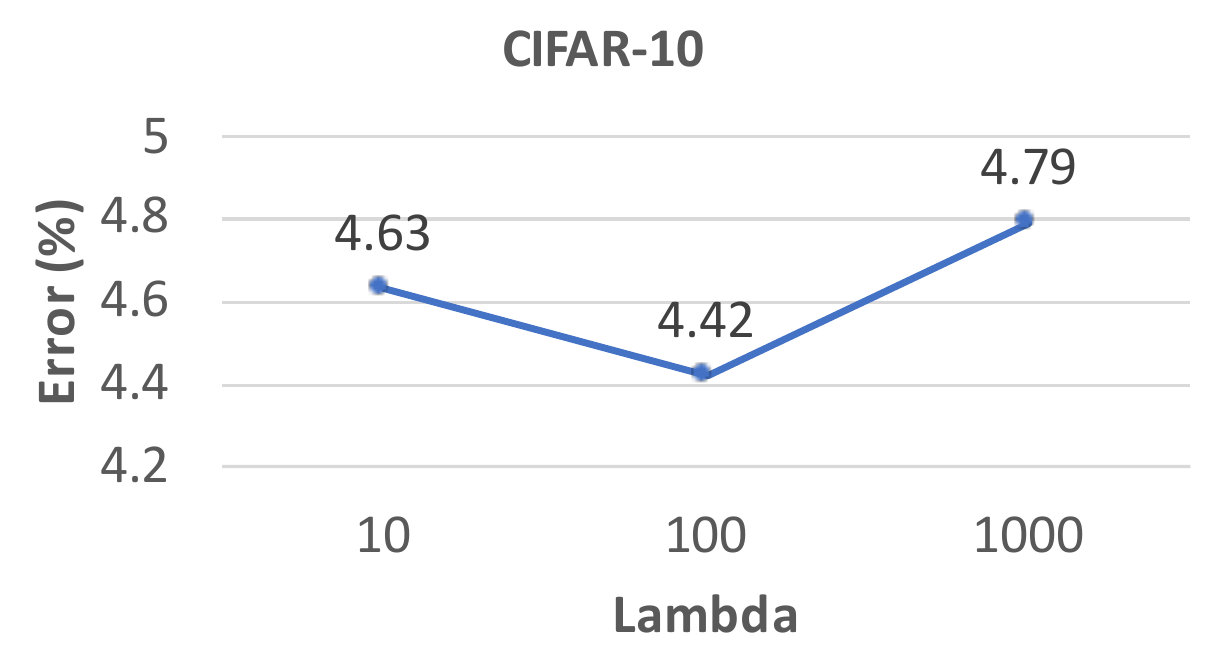}
       \caption{How classification errors on CIFAR-100 and CIFAR-10 change as $\lambda$ increases.}
 \label{fig:lambda}
\end{figure}

\begin{itemize}[leftmargin=*]
    \item Ablation study on the tradeoff parameter $\lambda$. We explore how the learners' performance varies as the tradeoff parameter $\lambda$ in Eq.(\ref{eq:reg}) increases.  For both CIFAR-100 and CIFAR-10, we randomly sample 5K data from the 25K training  and  25K validation data, and use it as a test set to report performance in this ablation study. 
    The rest 45K data (22.5K training data and 22.5K validation data) is used for architecture search and evaluation. IL is applied to DARTS-2nd. The number of rounds is set to 2. 
    \item Ablation study on the number of rounds. In this study, we explore how the test error changes as we increase the number of interleaving rounds $M$ from 1 to 3. The results are reported on the 5K sampled data. 
    In this experiment, the tradeoff parameter $\lambda$ is set to 100. IL is applied to DRATS-2nd. 
    \item Ablation study on the order of tasks. In this study, we explore whether the order of tasks affects the test error. We experimented two orders (with the number of rounds set to 2): 1) CIFAR-100, CIFAR-10, CIFAR-100, CIFAR-10; 2) CIFAR-10, CIFAR-100, CIFAR-10, CIFAR-100. In order 1, classification on CIFAR-100 is performed first; in order 2, classification on CIFAR-10 is performed first. 
    In this experiment, the tradeoff parameter $\lambda$ is set to 100.
\end{itemize}

\begin{figure}[t]
    \centering
 \includegraphics[width=0.49\columnwidth]{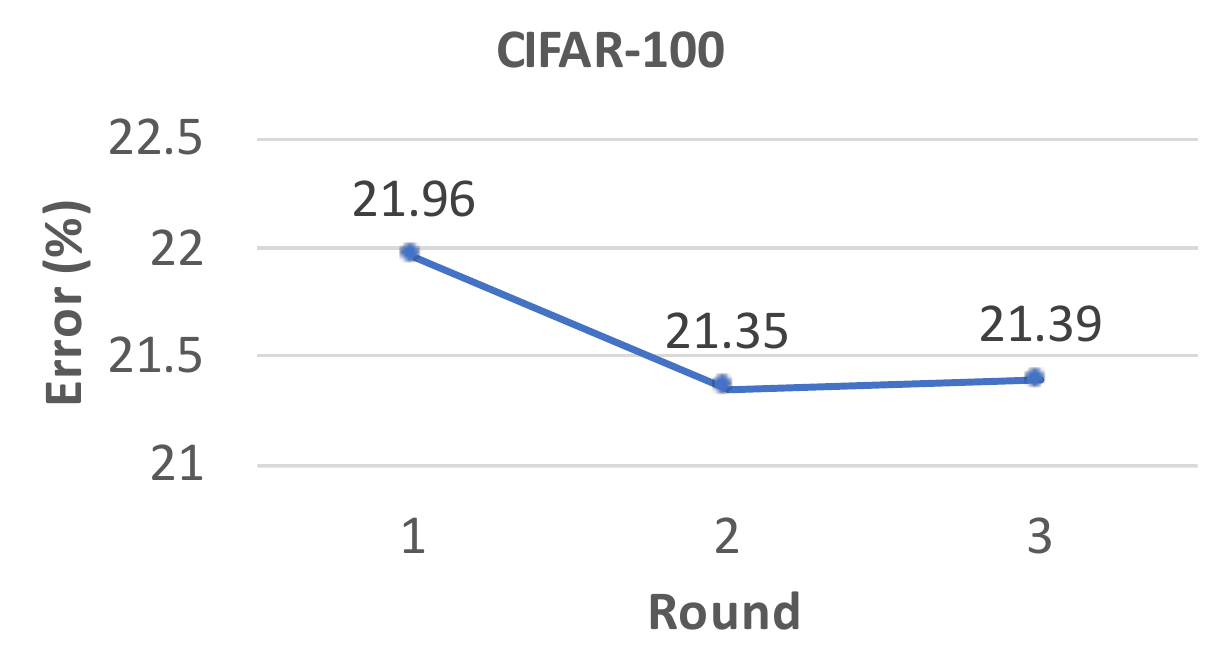}
  \includegraphics[width=0.49\columnwidth]{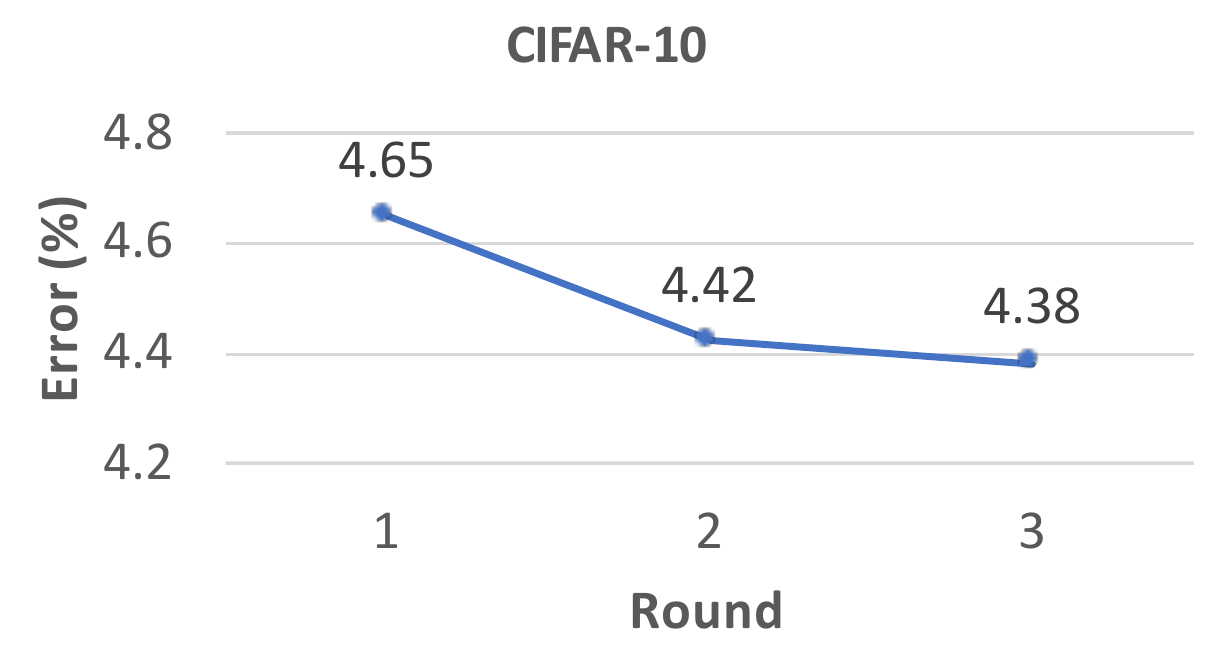}
       \caption{How classification errors on CIFAR-100 and CIFAR-10 change as the number of interleaving rounds $M$ increases.}
 \label{fig:round}
\end{figure}

\begin{table}[t]
\caption{Results for ablation study on the order of tasks. ``Order 1" denotes ``CIFRA-100, CIFAR-10, CIFAR-100, CIFAR-10". ``Order 2" denotes ``CIFRA-10, CIFAR-100, CIFAR-10, CIFAR-100". 
    }
    \centering
    \begin{tabular}{l|c}
    \hline
    Method & Error (\%)\\
    \hline
   Order 1 (CIFAR-100) &  17.12$\pm$0.08 \\ 
            Order 2  (CIFAR-100) &17.19$\pm$0.14  \\
         \hline
             Order 1 (CIFAR-10) & 2.73$\pm$0.04  \\
         Order 2  (CIFAR-10) &2.79$\pm$0.11   \\
         \hline
    \end{tabular}
    \label{tab:order}
\end{table}

Figure~\ref{fig:lambda} shows how the classification errors on the test sets of CIFAR-100 and CIFAR-10 vary as the tradeoff parameter $\lambda$ increases. As can be seen, for both datasets, when $\lambda$ increases from  10 to 100, the errors decrease. A larger $\lambda$ encourages a stronger knowledge transfer effect: the learning of the current learner C is sufficiently influenced by the previous learner P; the well-trained data encoder of P can effectively help to train the encoder of C, which results in better classification performance. However, further increasing  $\lambda$ renders the errors to increase. This is because an excessively large $\lambda$ will make the encoder of C strongly biased to the encoder of P while ignoring the specific data patterns in C's own training data. Since P's encoder may not be suitable for representing C's data, such a bias leads to inferior classification performance.

Figure~\ref{fig:round} shows how the classification errors on the test sets of CIFAR-100 and CIFAR-10 vary as the number of rounds $M$ increases. For CIFAR-100, when $M$ increases from 1 to 2, the error is reduced. When $M=1$, the interleaving effect is weak:  classification on CIFAR-100 influences classification on CIFAR-10, but not the other way around. When $M=2$, the interleaving effect is strong: CIFAR-100 influences CIFAR-10 and CIFAR-10 in turn influences CIFAR-100. This further demonstrates the effectiveness of interleaving learning.
Increasing $M$ from 2 to 3 does not significantly reduce the error further. This is probably because 2 rounds of interleaving have brought in sufficient interleaving effect. Similar trend is observed in the plot of CIFAR-10.

Table~\ref{tab:order} shows the test errors on CIFAR-100 and CIFAR-10 under two different orders. In order 1, the starting task is classification on CIFAR-100. In order 2, the starting task is classification on CIFAR-10. As can be seen, the errors are not affected by the task order significantly. The reason is that: via interleaving, each task influences the other task at some point in the interleaving sequence; therefore, it does not matter too much regarding which task should be performed first.

\section{Conclusions and Future Works}
In this paper, we propose a novel machine learning framework  called  interleaving learning (IL). In IL, multiple tasks are performed in an interleaving fashion where task 1 is performed for a short while, then task 2 is conducted, then task 3, etc. After all tasks are learned in one round, the learning goes back to task 1 and the cyclic procedure starts over. These tasks share a data encoder, whose network weights are trained successively by different tasks in the interleaving process.  Via interleaving, different models transfer their learned knowledge to each other to better represent data and avoid being stuck in bad local optimums. We propose a multi-level optimization framework to formulate interleaving learning, where different learning stages are performed end-to-end. An efficient gradient-based algorithm is developed to solve the multi-level optimization problem. 
Experiments of neural architecture search  on  CIFAR-100 and CIFAR-10  demonstrate the effectiveness of interleaving learning. 

For future works, we will investigate other mechanisms that enable adjacent learners in the interleaving sequence to transfer knowledge, such as based on pseudo-labeling or self-supervised learning.

\bibliography{release}

\begin{thebibliography}{35}
\providecommand{\natexlab}[1]{#1}
\providecommand{\url}[1]{\texttt{#1}}
\expandafter\ifx\csname urlstyle\endcsname\relax
  \providecommand{\doi}[1]{doi: #1}\else
  \providecommand{\doi}{doi: \begingroup \urlstyle{rm}\Url}\fi

\bibitem[Cai et~al.(2019)Cai, Zhu, and Han]{cai2018proxylessnas}
Han Cai, Ligeng Zhu, and Song Han.
\newblock Proxylessnas: Direct neural architecture search on target task and
  hardware.
\newblock In \emph{{ICLR}}, 2019.

\bibitem[Casale et~al.(2019)Casale, Gordon, and Fusi]{abs-1902-05116}
Francesco~Paolo Casale, Jonathan Gordon, and Nicol{\'{o}} Fusi.
\newblock Probabilistic neural architecture search.
\newblock \emph{CoRR}, abs/1902.05116, 2019.

\bibitem[Chen and Hsieh(2020)]{abs-2002-05283}
Xiangning Chen and Cho{-}Jui Hsieh.
\newblock Stabilizing differentiable architecture search via perturbation-based
  regularization.
\newblock \emph{CoRR}, abs/2002.05283, 2020.

\bibitem[Chen et~al.(2020)Chen, Wang, Cheng, Tang, and Hsieh]{abs-2006-10355}
Xiangning Chen, Ruochen Wang, Minhao Cheng, Xiaocheng Tang, and Cho{-}Jui
  Hsieh.
\newblock Drnas: Dirichlet neural architecture search.
\newblock \emph{CoRR}, abs/2006.10355, 2020.

\bibitem[Chu et~al.(2019)Chu, Zhou, Zhang, and Li]{abs-1911-12126}
Xiangxiang Chu, Tianbao Zhou, Bo~Zhang, and Jixiang Li.
\newblock Fair {DARTS:} eliminating unfair advantages in differentiable
  architecture search.
\newblock \emph{CoRR}, abs/1911.12126, 2019.

\bibitem[Chu et~al.(2020{\natexlab{a}})Chu, Wang, Zhang, Lu, Wei, and
  Yan]{abs-2009-01027}
Xiangxiang Chu, Xiaoxing Wang, Bo~Zhang, Shun Lu, Xiaolin Wei, and Junchi Yan.
\newblock {DARTS-:} robustly stepping out of performance collapse without
  indicators.
\newblock \emph{CoRR}, abs/2009.01027, 2020{\natexlab{a}}.

\bibitem[Chu et~al.(2020{\natexlab{b}})Chu, Zhang, and Li]{abs-2005-03566}
Xiangxiang Chu, Bo~Zhang, and Xudong Li.
\newblock Noisy differentiable architecture search.
\newblock \emph{CoRR}, abs/2005.03566, 2020{\natexlab{b}}.

\bibitem[Deng et~al.(2009)Deng, Dong, Socher, Li, Li, and
  Fei-Fei]{deng2009imagenet}
Jia Deng, Wei Dong, Richard Socher, Li-Jia Li, Kai Li, and Li~Fei-Fei.
\newblock Imagenet: A large-scale hierarchical image database.
\newblock In \emph{2009 IEEE conference on computer vision and pattern
  recognition}, pages 248--255. Ieee, 2009.

\bibitem[Dong and Yang(2019)]{DongY19}
Xuanyi Dong and Yi~Yang.
\newblock Searching for a robust neural architecture in four {GPU} hours.
\newblock In \emph{{CVPR}}, 2019.

\bibitem[He et~al.(2016)He, Zhang, Ren, and Sun]{he2016deep}
Kaiming He, Xiangyu Zhang, Shaoqing Ren, and Jian Sun.
\newblock Deep residual learning for image recognition.
\newblock In \emph{CVPR}, 2016.

\bibitem[Hong et~al.(2020)Hong, Li, Zhang, Tang, Wang, Li, and
  Yu]{HongL0TWL020}
Weijun Hong, Guilin Li, Weinan Zhang, Ruiming Tang, Yunhe Wang, Zhenguo Li, and
  Yong Yu.
\newblock Dropnas: Grouped operation dropout for differentiable architecture
  search.
\newblock In \emph{{IJCAI}}, 2020.

\bibitem[Howard et~al.(2017)Howard, Zhu, Chen, Kalenichenko, Wang, Weyand,
  Andreetto, and Adam]{HowardZCKWWAA17}
Andrew~G. Howard, Menglong Zhu, Bo~Chen, Dmitry Kalenichenko, Weijun Wang,
  Tobias Weyand, Marco Andreetto, and Hartwig Adam.
\newblock Mobilenets: Efficient convolutional neural networks for mobile vision
  applications.
\newblock \emph{CoRR}, abs/1704.04861, 2017.

\bibitem[Hu et~al.(2020)Hu, Xie, Zheng, Liu, Shi, Liu, and Lin]{HuXZLSLL20}
Shoukang Hu, Sirui Xie, Hehui Zheng, Chunxiao Liu, Jianping Shi, Xunying Liu,
  and Dahua Lin.
\newblock {DSNAS:} direct neural architecture search without parameter
  retraining.
\newblock In \emph{{CVPR}}, 2020.

\bibitem[Huang et~al.(2017)Huang, Liu, van~der Maaten, and
  Weinberger]{HuangLMW17}
Gao Huang, Zhuang Liu, Laurens van~der Maaten, and Kilian~Q. Weinberger.
\newblock Densely connected convolutional networks.
\newblock In \emph{{CVPR}}, 2017.

\bibitem[Kingma and Ba(2014)]{adam}
Diederik Kingma and Jimmy Ba.
\newblock Adam: A method for stochastic optimization.
\newblock \emph{International Conference on Learning Representations}, 12 2014.

\bibitem[Liang et~al.(2019)Liang, Zhang, Sun, He, Huang, Zhuang, and
  Li]{abs-1909-06035}
Hanwen Liang, Shifeng Zhang, Jiacheng Sun, Xingqiu He, Weiran Huang, Kechen
  Zhuang, and Zhenguo Li.
\newblock {DARTS+:} improved differentiable architecture search with early
  stopping.
\newblock \emph{CoRR}, abs/1909.06035, 2019.

\bibitem[Lin et~al.(2014)Lin, Maire, Belongie, Hays, Perona, Ramanan,
  Doll{\'a}r, and Zitnick]{coco}
Tsung-Yi Lin, Michael Maire, Serge Belongie, James Hays, Pietro Perona, Deva
  Ramanan, Piotr Doll{\'a}r, and C~Lawrence Zitnick.
\newblock Microsoft coco: Common objects in context.
\newblock In \emph{ECCV}, 2014.

\bibitem[Liu et~al.(2018{\natexlab{a}})Liu, Zoph, Neumann, Shlens, Hua, Li,
  Fei{-}Fei, Yuille, Huang, and Murphy]{LiuZNSHLFYHM18}
Chenxi Liu, Barret Zoph, Maxim Neumann, Jonathon Shlens, Wei Hua, Li{-}Jia Li,
  Li~Fei{-}Fei, Alan~L. Yuille, Jonathan Huang, and Kevin Murphy.
\newblock Progressive neural architecture search.
\newblock In \emph{{ECCV}}, 2018{\natexlab{a}}.

\bibitem[Liu et~al.(2018{\natexlab{b}})Liu, Simonyan, Vinyals, Fernando, and
  Kavukcuoglu]{liu2017hierarchical}
Hanxiao Liu, Karen Simonyan, Oriol Vinyals, Chrisantha Fernando, and Koray
  Kavukcuoglu.
\newblock Hierarchical representations for efficient architecture search.
\newblock In \emph{{ICLR}}, 2018{\natexlab{b}}.

\bibitem[Liu et~al.(2019)Liu, Simonyan, and Yang]{liu2018darts}
Hanxiao Liu, Karen Simonyan, and Yiming Yang.
\newblock {DARTS:} differentiable architecture search.
\newblock In \emph{{ICLR}}, 2019.

\bibitem[Luo et~al.(2018)Luo, Tian, Qin, Chen, and Liu]{LuoTQCL18}
Renqian Luo, Fei Tian, Tao Qin, Enhong Chen, and Tie{-}Yan Liu.
\newblock Neural architecture optimization.
\newblock In \emph{NeurIPS}, 2018.

\bibitem[Ma et~al.(2018)Ma, Zhang, Zheng, and Sun]{MaZZS18}
Ningning Ma, Xiangyu Zhang, Hai{-}Tao Zheng, and Jian Sun.
\newblock Shufflenet {V2:} practical guidelines for efficient {CNN}
  architecture design.
\newblock In \emph{{ECCV}}, 2018.

\bibitem[Noy et~al.(2020)Noy, Nayman, Ridnik, Zamir, Doveh, Friedman, Giryes,
  and Zelnik]{NoyNRZDFGZ20}
Asaf Noy, Niv Nayman, Tal Ridnik, Nadav Zamir, Sivan Doveh, Itamar Friedman,
  Raja Giryes, and Lihi Zelnik.
\newblock {ASAP:} architecture search, anneal and prune.
\newblock In \emph{{AISTATS}}, 2020.

\bibitem[Pham et~al.(2018)Pham, Guan, Zoph, Le, and Dean]{pham2018efficient}
Hieu Pham, Melody~Y. Guan, Barret Zoph, Quoc~V. Le, and Jeff Dean.
\newblock Efficient neural architecture search via parameter sharing.
\newblock In \emph{{ICML}}, 2018.

\bibitem[Real et~al.(2019)Real, Aggarwal, Huang, and Le]{real2019regularized}
Esteban Real, Alok Aggarwal, Yanping Huang, and Quoc~V Le.
\newblock Regularized evolution for image classifier architecture search.
\newblock In \emph{Proceedings of the aaai conference on artificial
  intelligence}, volume~33, pages 4780--4789, 2019.

\bibitem[Szegedy et~al.(2015)Szegedy, Liu, Jia, Sermanet, Reed, Anguelov,
  Erhan, Vanhoucke, and Rabinovich]{googlenet}
Christian Szegedy, Wei Liu, Yangqing Jia, Pierre Sermanet, Scott Reed, Dragomir
  Anguelov, Dumitru Erhan, Vincent Vanhoucke, and Andrew Rabinovich.
\newblock Going deeper with convolutions.
\newblock In \emph{CVPR}, 2015.

\bibitem[Tan et~al.(2019)Tan, Chen, Pang, Vasudevan, Sandler, Howard, and
  Le]{TanCPVSHL19}
Mingxing Tan, Bo~Chen, Ruoming Pang, Vijay Vasudevan, Mark Sandler, Andrew
  Howard, and Quoc~V. Le.
\newblock Mnasnet: Platform-aware neural architecture search for mobile.
\newblock In \emph{{CVPR}}, 2019.

\bibitem[Wang et~al.(2020)Wang, Xue, Yan, Yang, Hu, and Sun]{WangXYYHS20}
Xiaoxing Wang, Chao Xue, Junchi Yan, Xiaokang Yang, Yonggang Hu, and Kewei Sun.
\newblock Mergenas: Merge operations into one for differentiable architecture
  search.
\newblock In \emph{{IJCAI}}, 2020.

\bibitem[Xie et~al.(2019)Xie, Zheng, Liu, and Lin]{xie2018snas}
Sirui Xie, Hehui Zheng, Chunxiao Liu, and Liang Lin.
\newblock {SNAS:} stochastic neural architecture search.
\newblock In \emph{{ICLR}}, 2019.

\bibitem[Xu et~al.(2020)Xu, Xie, Zhang, Chen, Qi, Tian, and
  Xiong]{abs-1907-05737}
Yuhui Xu, Lingxi Xie, Xiaopeng Zhang, Xin Chen, Guo{-}Jun Qi, Qi~Tian, and
  Hongkai Xiong.
\newblock {PC-DARTS:} partial channel connections for memory-efficient
  architecture search.
\newblock In \emph{{ICLR}}, 2020.

\bibitem[Zela et~al.(2020)Zela, Elsken, Saikia, Marrakchi, Brox, and
  Hutter]{ZelaESMBH20}
Arber Zela, Thomas Elsken, Tonmoy Saikia, Yassine Marrakchi, Thomas Brox, and
  Frank Hutter.
\newblock Understanding and robustifying differentiable architecture search.
\newblock In \emph{{ICLR}}, 2020.

\bibitem[Zhang et~al.(2018)Zhang, Zhou, Lin, and Sun]{ZhangZLS18}
Xiangyu Zhang, Xinyu Zhou, Mengxiao Lin, and Jian Sun.
\newblock Shufflenet: An extremely efficient convolutional neural network for
  mobile devices.
\newblock In \emph{{CVPR}}, 2018.

\bibitem[Zhou et~al.(2019)Zhou, Yang, Wang, and Pan]{ZhouYWP19}
Hongpeng Zhou, Minghao Yang, Jun Wang, and Wei Pan.
\newblock Bayesnas: {A} bayesian approach for neural architecture search.
\newblock In \emph{{ICML}}, 2019.

\bibitem[Zoph and Le(2017)]{zoph2016neural}
Barret Zoph and Quoc~V. Le.
\newblock Neural architecture search with reinforcement learning.
\newblock In \emph{{ICLR}}, 2017.

\bibitem[Zoph et~al.(2018)Zoph, Vasudevan, Shlens, and Le]{zoph2018learning}
Barret Zoph, Vijay Vasudevan, Jonathon Shlens, and Quoc~V Le.
\newblock Learning transferable architectures for scalable image recognition.
\newblock In \emph{CVPR}, 2018.

\end{thebibliography}

\end{document}